\documentclass[letterpaper, 10 pt, conference]{ieeeconf}  %
\IEEEoverridecommandlockouts
\usepackage{cite}
\usepackage{amsmath,amssymb,amsfonts}
\usepackage{algorithmic}
\usepackage{graphicx}
\usepackage{siunitx}
\usepackage{comment}
\usepackage{hyperref}
\usepackage{textcomp}
\def\BibTeX{{\rm B\kern-.05em{\sc i\kern-.025em b}\kern-.08em
    T\kern-.1667em\lower.7ex\hbox{E}\kern-.125emX}}
\usepackage{censor}
\newcommand{\methodeName}[1][RUMI]{#1}
\usepackage{xcolor}

\usepackage{booktabs,tabularx,array,colortbl}

\definecolor{rwHighlight}{RGB}{235,244,243}

\usepackage{float}

\newcolumntype{K}{>{\centering\arraybackslash}X}

\usepackage{booktabs,tabularx,array,amssymb,colortbl}
\definecolor{rumiHybridHighlight}{RGB}{235,243,244}

\usepackage{hyperref}
\usepackage[caption=false,font=footnotesize]{subfig}
\definecolor{rumiCompactHighlight}{RGB}{235,243,244}

\providecommand{\rcYes}{\ensuremath{\checkmark}}
\providecommand{\rcNo}{\textendash}

\providecommand{\rcFeature}[1]{%
    \mbox{\rcYes\,#1}%
}

\providecommand{\rcHead}[2]{%
    \begin{tabular}[b]{@{}l@{}}
        \textbf{#1}\\
        \textbf{#2}
    \end{tabular}%
}

\StopCensoring
\title{ \LARGE \bf
\methodeName: A Robotiq-Matched Universal Manipulation Interface for Robot Learning with Augmented Reality and Haptic Feedback}

\author{\censor{Kevin Avdiu$^{*1,3}$, Seongjin Bien$^{*2}$, Yannik Blei$^{*2}$, Philine Meister$^{1}$,\\
Johannes Hechtl$^{1,2}$, Philipp Schmitt$^{1}$, Kai Wurm$^{1}$, Georg von Wichert$^{1}$, Wolfram Burgard$^{*2}$}
\thanks{\censor{$^*$Co-authors, sorted by last name. Contact: yannik.blei@utn.de}}%
\thanks{\blackout{$^{1}$Siemens, Germany. }}%
\thanks{\blackout{$^{2}$University of Technology Nuremberg, Germany.}}
\thanks{\blackout{$^{3}$Technical University of Munich, Germany.}}
}

\begin{document}
\bstctlcite{IEEEexample:BSTcontrol}
\maketitle

\suppressfloats[t]

\begin{figure}[t]
    \centering
    \includegraphics[width=.95\linewidth]
        {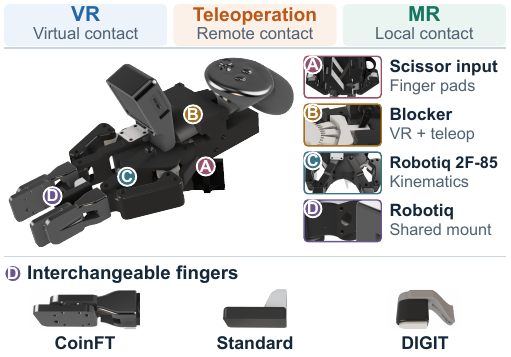}
    \caption{RUMI supports three demonstration modes: virtual reality (VR), teleoperation, and mixed reality (MR). The hardware features scissor input, selective blocking, Robotiq-matched kinematics, and interchangeable fingers, including CoinFT, standard, and DIGIT.}
    \label{fig:rumi_teaser}
    \vspace{-1em}
\end{figure}

\thispagestyle{empty}

    \begin{abstract}
Many manipulation tasks require carefully controlled contact forces for successful execution, yet large-scale, force-aware datasets remain scarce.
Handheld interfaces can help by transmitting interaction forces directly through the operator's arm, but most existing designs are tailored to simple linear grippers and do not match the nonlinear mechanics of widely used robotic grippers. We present \methodeName{}, a handheld universal manipulation interface matched to the Robotiq 2F-85 with exchangeable fingertip sensors that preserves the target gripper geometry while enabling natural input before contact and force measurement after contact. By integrating the tracking system into a virtual reality headset, our method provides real-time augmented reality feedback about the kinematic feasibility of motion on the target robot platform.
We further introduce a haptic feedback mechanism that supports force-aware teleoperation for simulated and real robots. A user study shows fewer episodes containing joint-limit violations with live robot visualization and pose initialization (3/50) than without augmented reality (16/50). We additionally train imitation policies from demonstrations, collected using two different fingertip modalities enabled by the gripper-matched hardware, and assess the haptic interface's capability for collecting force-aware teleoperation datasets in simulation. The code and assets for RUMI are available at: https://utn-air.github.io/RUMI/.
\end{abstract}

\section{Introduction}

Learning from demonstrations reduces the effort required to program robots for complex or variable tasks~\cite{argall2009survey,liu2024arprogramming}. This is increasingly relevant for large visuomotor policies, whose performance depends on the scale and quality of their demonstration data~\cite{kim2024openvlaopensourcevisionlanguageactionmodel}. 
Collaborative efforts such as Open X-Embodiment~\cite{openx2024} aggregate standardized manipulation data across robot platforms, tasks, and institutions using traditional teleoperation approaches, which are typically time-consuming. This limitation has motivated the creation of handheld data collection devices that enable rapid data collection. Universal Manipulation Interfaces (UMIs) support portable, robot-free data collection and thereby decouple data acquisition from continuous access to the target robot~\cite{chi2024universalmanipulationinterfaceinthewild,zhaxizhuoma2025fastumiscalablehardwareindependentuniversal,liu2025fastumi100kadvancingdatadrivenrobotic}. These devices have also gained popularity in the tactile manipulation community, where collecting large-scale datasets remains challenging due to the inherent sparsity of tactile sensor signals~\cite{choi2026inthewildcompliantmanipulationumift,farm2026,wu2025freetacman}. 

Despite the scalability and speed-up they offer, the inherent robot-free nature of these approaches means that it becomes difficult to observe the validity of the trajectories which they generate. While an intuition for robot-compatible motion can be learned through practice, the lack of feedback necessitates careful post-processing to filter out any unusable data. Furthermore, most existing methods are designed around custom-built robotic grippers, which makes them unsuitable for applications that require more robust grippers.

Another underexplored potential of UMI devices is that they can serve as a more intuitive interface for teleoperation, given their similar appearances and kinematics to target grippers. However, most existing UMI designs lack a way to transmit contact information at time of grasp, which makes it difficult to regulate grasp forces in such scenarios.

To address these challenges, we present the Robotiq Universal Manipulation Interface (RUMI), a handheld platform matched to the nonlinear kinematics, finger attachments, and wrist-camera viewpoint of the Robotiq 2F-85. A scissor-like input drives the coupled fingers, while a rotary encoder measures the resulting grasp coordinate. Interchangeable fingers support force/torque and vision-based tactile sensing for contact-rich demonstrations. During simulated and real-robot teleoperation, a contact-triggered blocker provides low-resistance aperture control before contact and converts the subsequent squeeze into a measured force command. By integrating the handheld pose tracking into a virtual reality (VR) headset, we can project a digital shadow of the target robot platform in augmented reality (AR) fashion during data collection, allowing the operator to recognize infeasible poses while recording rather than filtering them only after collection.

\noindent \textbf{In summary, we make the following contributions:}
\begin{itemize}
    \item RUMI, a Robotiq 2F-85-matched handheld interface with nonlinear actuation and interchangeable fingertips;
    \item An AR pipeline for visualizing and validating robot configurations during robot-free data collection;
    \item A haptic feedback mechanism which allows for gripper force control during simulated and real teleoperation;
    \item Releasing the device design and software framework as well as a dataset of demonstrations featuring force/torque and vision-based tactile sensor data.
\end{itemize}

\section{Related Work}

\begin{table*}[t!]
\vspace{5.1pt}
\caption{Reported capabilities of handheld demonstration interfaces. \rcYes: reported; \rcNo: not reported. Grasp force excludes wrist-only loads; visual guidance is operator-facing; added haptics excludes natural contact.}
\label{tab:related_work_compact}
\centering
\begingroup
\footnotesize
\setlength{\tabcolsep}{2pt}
\renewcommand{\arraystretch}{0.97}
\begin{tabularx}{\textwidth}{>{\raggedright\arraybackslash}p{84pt}>{\raggedright\arraybackslash}p{63pt}>{\raggedright\arraybackslash}p{61pt}>{\raggedright\arraybackslash}p{55pt}>{\raggedright\arraybackslash}p{49pt}>{\raggedright\arraybackslash}p{78pt}>{\raggedright\arraybackslash}X}
\toprule
\textbf{Interface} & \rcHead{Contact}{sensing} & \rcHead{Grasp}{force} & \rcHead{Live robot}{visualization} & \rcHead{Gripper}{haptics} & \rcHead{Tracking /}{capture setup} & \rcHead{Robot}{counterpart} \\
\midrule
\multicolumn{7}{@{}l}{\textit{Two-finger interfaces without fingertip contact sensors}} \\
\mbox{UMI~\cite{chi2024universalmanipulationinterfaceinthewild}} & \mbox{\rcNo} & \mbox{\rcNo} & \mbox{\rcNo} & \mbox{\rcNo} & \mbox{Visuoinertial SLAM} & \mbox{WSG-50} \\
\mbox{FastUMI~\cite{zhaxizhuoma2025fastumiscalablehardwareindependentuniversal}} & \mbox{\rcNo} & \mbox{\rcNo} & \mbox{\rcNo} & \mbox{\rcNo} & \mbox{Realsense T265} & \mbox{Parallel gripper} \\
\mbox{ForceMimic~\cite{liu2024forcemimic}} & \mbox{Wrist force} & \mbox{\rcNo} & \mbox{\rcNo} & \mbox{\rcNo} & \mbox{Realsense T265} & \mbox{Parallel gripper} \\
\mbox{ManipForce~\cite{lee2025manipforce}} & \mbox{Wrist force} & \mbox{\rcNo} & \mbox{\rcNo} & \mbox{\rcNo} & \mbox{ArUco tracking} & \mbox{Parallel gripper} \\
\mbox{ActiveUMI~\cite{zeng2025activeumi}} & \mbox{\rcNo} & \mbox{\rcNo} & \mbox{\rcFeature{AR Projection}} & \mbox{\rcNo} & \mbox{Quest VR + Backpack} & \mbox{Custom bimanual system} \\
\mbox{UMIGen~\cite{huang2025umigen}} & \mbox{\rcNo} & \mbox{\rcNo} & \mbox{\rcNo} & \mbox{\rcNo} & \mbox{Realsense T265} & \mbox{Parallel gripper} \\
\mbox{XRZero-G0~\cite{wang2026xrzero}} & \mbox{\rcNo} & \mbox{\rcNo} & \mbox{\rcNo} & \mbox{\rcNo} & \mbox{PICO VR + backpack} & \mbox{Pinch gripper} \\
\mbox{UMI-3D~\cite{wang2026umi3d}} & \mbox{\rcNo} & \mbox{\rcNo} & \mbox{\rcNo} & \mbox{\rcNo} & \mbox{Onboard LiDAR} & \mbox{Parallel gripper} \\
\mbox{YUBI~\cite{ohkawa2026yubiyieldinguniversalbidigital}} & \mbox{\rcNo} & \mbox{\rcNo} & \mbox{\rcNo} & \mbox{\rcNo} & \mbox{Static Quest VR} & \mbox{Pinch gripper} \\
\mbox{EgoGuide~\cite{xu2026egoguide}} & \mbox{\rcNo} & \mbox{\rcNo} & \mbox{\rcNo} & \mbox{\rcNo} & \mbox{Quest VR} & \mbox{Flexiv Grav} \\
\mbox{HiFi-UMI~\cite{ai2026hifiumilearningdeployablemanipulation}} & \mbox{\rcNo} & \mbox{\rcNo} & \mbox{\rcNo} & \mbox{\rcNo} & \mbox{Offl. SLAM + ArUco} & \mbox{Pinch gripper} \\
\addlinespace[1.5pt]
\multicolumn{7}{@{}l}{\textit{Two-finger interfaces with fingertip contact sensors}} \\
\mbox{ViTaMIn~\cite{liu2025vitaminlearningcontactrichtasks}} & \mbox{Optical tactile} & \mbox{\rcNo} & \mbox{\rcNo} & \mbox{\rcNo} & \mbox{Visuoinertial SLAM} & \mbox{PGI-140-80-W-S} \\
\mbox{FreeTacMan~\cite{wu2025freetacman}} & \mbox{Optical tactile} & \mbox{\rcNo} & \mbox{\rcNo} & \mbox{\rcNo} & \mbox{NOKOV} & \mbox{Parallel gripper} \\
\mbox{Touch in the Wild~\cite{zhu2025touchwild}} & \mbox{Resistive tactile} & \mbox{\rcNo} & \mbox{\rcNo} & \mbox{\rcNo} & \mbox{Visuoinertial SLAM} & \mbox{Parallel gripper} \\
\mbox{exUMI~\cite{xu2025exumi}} & \mbox{Modular tactile} & \mbox{\rcNo} & \mbox{\rcNo} & \mbox{\rcNo} & \mbox{Quest} & \mbox{Grav / Robotiq 2F} \\
\mbox{MagiClaw~\cite{wu2025magiclaw}} & \mbox{Optical tactile} & \mbox{\rcFeature{Estimated}} & \mbox{\rcNo} & \mbox{\rcFeature{Motors}} & \mbox{iPhone SLAM} & \mbox{Shared MagiClaw} \\
\mbox{FARM~\cite{farm2026}} & \mbox{Optical tactile} & \mbox{\rcFeature{Estimated}} & \mbox{\rcNo} & \mbox{\rcNo} & \mbox{OptiTrack} & \mbox{Actuated UMI} \\
\mbox{ViTaMIn-B~\cite{li2025vitaminb}} & \mbox{Optical tactile} & \mbox{\rcNo} & \mbox{\rcNo} & \mbox{\rcNo} & \mbox{Quest} & \mbox{Parallel gripper} \\
\mbox{UMI-FT~\cite{choi2026inthewildcompliantmanipulationumift}} & \mbox{Finger force} & \mbox{\rcFeature{Direct}} & \mbox{\rcNo} & \mbox{\rcNo} & \mbox{iPhone SLAM} & \mbox{WSG-50} \\
\mbox{TacUMI~\cite{cheng2026tacumimultimodaluniversalmanipulation}} & \mbox{Tactile + wrist} & \mbox{\rcNo} & \mbox{\rcNo} & \mbox{\rcNo} & \mbox{Vive tracking stations} & \mbox{Parallel gripper} \\
\mbox{OmniUMI~\cite{luo2026omniumiphysicallygroundedrobot}} & \mbox{Tactile + wrist} & \mbox{\rcFeature{Estimated}} & \mbox{\rcNo} & \mbox{\rcFeature{Motors}} & \mbox{Vive tracking stations} & \mbox{Parallel gripper} \\
\addlinespace[1.5pt]
\multicolumn{7}{@{}l}{\textit{Multi-finger grippers and dexterous hands}} \\
\mbox{DexUMI~\cite{xu2025dexumi}} & \mbox{Tactile arrays} & \mbox{\rcFeature{Distributed}} & \mbox{\rcNo} & \mbox{\rcNo} & \mbox{iPhone SLAM} & \mbox{Custom hand} \\
\mbox{MV-UMI~\cite{rayyan2025mvumiscalablemultiviewinterface}} & \mbox{\rcNo} & \mbox{\rcNo} & \mbox{\rcNo} & \mbox{\rcNo} & \mbox{GoPro + overhead} & \mbox{Custom hand} \\
\mbox{RealDexUMI~\cite{xu2026realdexumi}} & \mbox{Resistive tactile} & \mbox{\rcNo} & \mbox{\rcNo} & \mbox{\rcNo} & \mbox{Pose tracker} & \mbox{Custom hand} \\
\mbox{Koala Gripper~\cite{hajjahmad2026koalagripper}} & \mbox{\rcNo} & \mbox{\rcNo} & \mbox{\rcNo} & \mbox{\rcNo} & \mbox{Pose camera} & \mbox{Custom gripper} \\
\midrule
\rowcolor{rumiCompactHighlight}
\mbox{\textbf{RUMI (ours)}} & \mbox{Force / optical tac.} & \mbox{\rcFeature{Direct}} & \mbox{\rcFeature{Robot}} & \mbox{\rcFeature{Blocker}} & \mbox{Quest} & \mbox{Robotiq 2F-85~\cite{robotiq_2f85_product}} \\
\bottomrule
\end{tabularx}
\endgroup
\end{table*}

\subsection{Robot-Free Demonstration and Embodiment Transfer}

We provide a comprehensive overview of representative robot-free data collection approaches in Tab. I. Existing UMI-style systems use portable handheld grippers with gripper-centric observations and pose tracking to collect demonstrations without an active target robot ~\cite{chi2024universalmanipulationinterfaceinthewild,zhaxizhuoma2025fastumiscalablehardwareindependentuniversal,liu2025fastumi100kadvancingdatadrivenrobotic,ohkawa2026yubiyieldinguniversalbidigital,rayyan2025mvumiscalablemultiviewinterface}. The original UMI design already supports transfer across parallel grippers through replaceable finger geometries, while later systems explore closer correspondence to specific robot embodiments, such as exUMI~\cite{xu2025exumi}. ActiveUMI~\cite{zeng2025activeumi} further incorporates AR-based robot visualization during robot-free collection.

RUMI follows these directions by matching the nonlinear finger kinematics and mounting geometry of the Robotiq 2F-85, enabling straightforward deployment on existing robot platforms without requiring a custom robot-side gripper or additional compatibility solution. It also uses live robot projection during collection for online kinematic checking.

\subsection{Multimodal Handheld Devices and Haptic Teleoperation}

Several UMI systems extend pose and gripper-width measurements with tactile or force sensing for contact-rich manipulation~\cite{choi2026inthewildcompliantmanipulationumift,farm2026,li2025vitaminb,cheng2026tacumimultimodaluniversalmanipulation}. RUMI similarly supports interchangeable fingertip sensors through a compatible finger mount, allowing the sensing configuration to be adapted without changing the underlying gripper embodiment.

Haptic feedback and teleoperation have also been incorporated into UMI-style devices, including MagiClaw~\cite{wu2025magiclaw} and OmniUMI~\cite{luo2026omniumiphysicallygroundedrobot}. RUMI follows this direction with a selectively engaged gripper-haptic mechanism, allowing the same handheld device to support both in-the-wild collection and teleoperation.

Overall, prior work has explored robot-free collection, transferable gripper designs, tactile and force sensing, live robot visualization, and haptic teleoperation in different combinations. RUMI brings these capabilities together in a single handheld interface with direct correspondence to an existing commercial gripper.

\section{Method}
\subsection{Design Requirements}
RUMI is designed for industrial manipulation settings, where robots equipped with commercial grippers are commonly used for task automation. We derive four main requirements from this setting.

First, the interface should reproduce the geometry and grasp actuation of a commonly used industrial gripper, specifically the Robotiq 2F-85, such that demonstrations can be collected using an embodiment consistent with the target system. Second, the fingers should be modular to support different tactile sensing modalities depending on the task. Third, robot-free demonstrations should remain compatible with the target robot's workspace and kinematic constraints despite the robot being absent from the physical interaction loop. Finally, data collection should be intuitive and minimally intrusive to the production environment, while the same interface should also support remote teleoperation when direct access to the robot workspace is restricted due to safety or spatial constraints.

\subsection{Hardware and Sensing}
\begin{figure*}[!t]
\vspace{5.1pt}
    \centering
    \includegraphics[
        width=\textwidth,
        height=69.596mm,
        keepaspectratio
    ]{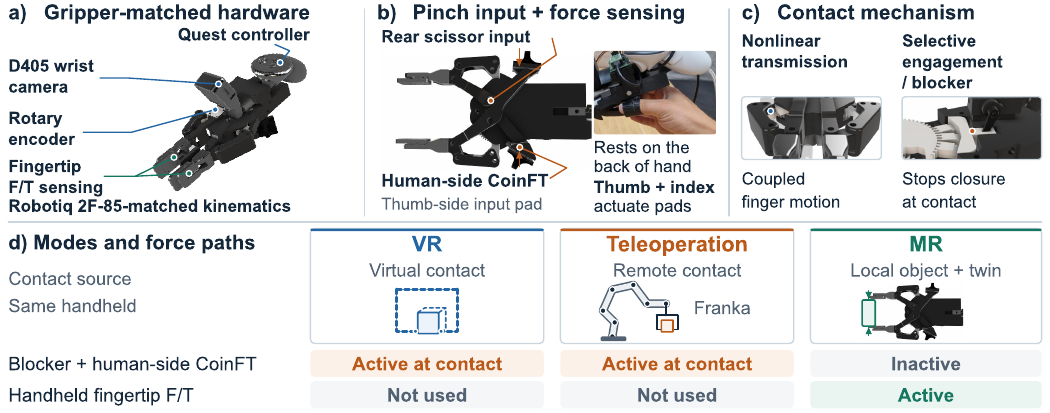}
    \caption{Overview of RUMI, a gripper-matched handheld interface
        for collecting force-aware manipulation demonstrations for robot learning.
        (a) The hardware combines Robotiq 2F-85-matched kinematics with
        a D405 wrist camera, a rotary encoder, a Quest controller,
        and fingertip force/torque (F/T) sensing.
        (b) The device rests on the back of the hand, while the thumb
        and index finger operate the rear scissor pads. A human-side
        CoinFT sensor measures force at the index-side input.
        (c) The nonlinear transmission couples finger motion, while
        a selectively engaged blocker constrains further closure.
        (d) The same interface supports three contact domains:
        virtual contact in VR, remote robot contact during teleoperation,
        and direct contact with a local object in MR.
        In VR and teleoperation, contact engages the blocker,
        providing resistance at the input pads; subsequent squeeze
        is measured by the human-side CoinFT sensor.
        In MR, the local object supplies the physical reaction force,
        and fingertip F/T sensors measure the interaction while
        the blocker and human-side force-sensing path remain inactive.
        Mode scenes are schematic.}
    \label{fig:rumi_overview_readability}
\end{figure*}

RUMI implements a crossed, scissor-like linkage that couples the motion of the operator's thumb and index finger to the gripper fingers (Fig.~\ref{fig:rumi_overview_readability}). Its linkage geometry and finger attachment interface are derived from the official Robotiq 2F-85 CAD model. The operator actuates two rear input pads, while an AS5600 magnetic rotary encoder measures the resulting gripper aperture. The scissor-like input interface allows direct mapping between operator's applied force to the gripper-side force, further complemented by a CoinFT~\cite{coinft} sensor on the index finger pad that captures the applied force.

The linkage can additionally be physically locked to provide haptic feedback (Fig.~\ref{fig:rumi_overview_readability}~c). A 9g Miuzei micro servo engages a toothed counterpart with the internal coupling gear, preventing further motion of the linkage. We selected this actuator because it is widely available and provides sufficient torque density for the compact mechanism.

The finger attachment points reproduce the 2F-85 mounting geometry, allowing the same finger modules to be transferred between RUMI and the robotic gripper, including the manufacturer's original fingers. We additionally provide CAD files for fingertip modules compatible with two sensing modalities: the CoinFT sensor, which measures a six-axis wrench at 200~Hz, and the DIGIT~\cite{digit} vision-based tactile sensor, which provides high-resolution tactile observations.

A Teensy 4.0 microcontroller coordinates the on-device sensing and actuation. It synchronizes measurements from the rotary encoder and any connected CoinFT sensors, transmits these measurements to the host computer via USB, and controls the servo used by the blocking mechanism. DIGIT sensors, when used, bypass the microcontroller and connect directly to the host computer.

A RealSense D405 is mounted at the wrist attachment point in the same pose used on the target robot. A Meta Quest 3 controller is rigidly attached to RUMI to provide six-DoF pose tracking relative to the headset.

 \subsection{Robot-State Visualization}

Robot-free data collection removes the target robot from the physical interaction loop and therefore also removes direct feedback about its workspace and kinematic limits. Without such feedback, an operator may generate motions that cannot be reproduced by the target robot.

To expose these constraints during demonstration, we extend the IRIS~\cite{iris2025jiang} AR application with support for RUMI. During recording, IRIS renders a MuJoCo-based~\cite{todorov2012mujoco} digital shadow of the target robot. The tracked RUMI pose is transformed into the robot base frame and used as the desired end-effector pose; an inverse-kinematics solver computes the corresponding robot configuration for visualization. The operator can therefore observe the robot configuration associated with the demonstrated motion in real time.

Providing meaningful feedback requires the virtual robot base to be accurately registered with the physical workspace. We use multiple QR codes to estimate the base pose and refine its orientation by fitting a plane to depth measurements of the deployment surface. This registration establishes the transformation between the headset tracking frame, the physical workspace, and the robot base frame. IRIS additionally supports configurable offsets from the detected base pose, allowing the virtual robot to be placed where its physical counterpart could not be accommodated because of spatial restrictions. 
An example is shown in Fig.~\ref{fig:setup}.

The same registration is used to express recorded RUMI trajectories in the robot base frame. Before starting an episode, the operator can additionally use a catch-to-teleop initialization procedure in which RUMI is aligned with the projected robot end-effector. This establishes a consistent initial end-effector pose while allowing the corresponding robot configuration to be inspected before and during the demonstration.

\subsection{Operating Modes and Contact Handling}

RUMI uses the same grasp interface for direct in-the-wild demonstration, simulated teleoperation, and real-robot teleoperation (Fig.~\ref{fig:rumi_overview_readability}). The modes differ primarily in how contact is generated and communicated to the operator.

During in-the-wild demonstrations, contact occurs directly between RUMI's fingers and the manipulated object. After the fingers make contact, the operator can continue squeezing the input pads, and the resulting force is measured by the actuation-side CoinFT sensor.

During teleoperation, contact occurs only at the remote or simulated gripper. Before contact, the blocking mechanism remains disengaged and the encoder measurement directly controls the commanded gripper aperture, allowing low-resistance motion of the RUMI linkage. When the remote gripper detects contact, the blocker engages and prevents further linkage closure. Subsequent squeezing is then measured by the actuation-side CoinFT sensor and mapped to a grasp-force command. An optional force amplification factor reduces the physical force required from the operator. When the operator releases the applied force, the blocker disengages and the interface returns to aperture-control mode.

The source of the contact signal depends on the teleoperation setting. In simulation, contact and force response are obtained from the simulated gripper; during real-robot teleoperation, contact is inferred from the physical gripper state and the measured RUMI force is sent as the gripper force command. The Meta Quest controller can additionally provide vibration feedback indicating remote contact and the approximate magnitude of the commanded load. This vibration complements the binary mechanical blocking cue but is not intended to reproduce an arbitrary external six-axis wrench.

\subsection{Data Recording and Policy Interface}

We use Robot Control Stack~\cite{juelg2026robotcontrolstack} as the underlying software architecture for data acquisition and robot communication. At each timestep, the system records the RUMI pose, gripper state, camera observations, and measurements from the selected fingertip sensors when available.

For robot-free demonstrations, RUMI poses are stored in the registered robot base frame established by the AR system. Robot actions can then be derived from the recorded trajectory either by temporally shifting the pose sequence or by computing the relative motion between consecutive poses.

During teleoperation, the RUMI input is instead used online as the commanded action, while the measured robot state is used as the observation. Both acquisition modes therefore share the same episode structure, while differing in whether actions are reconstructed from a robot-free demonstration or issued directly during teleoperation.

The resulting episodes can be converted into different observation-action representations for imitation learning. Observations can include camera streams, robot or RUMI poses, gripper state, blocking-mechanism state, and available fingertip-sensor measurements, while actions contain the desired robot motion and gripper command. For robot-free demonstrations, the inverse-kinematics solution associated with the AR projection can additionally provide joint configurations along the demonstrated trajectory, allowing the recordings to support both Cartesian pose-based and joint-space policy representations.

\section{Experimental Evaluation}

\begin{figure}[t]
    \centering
    \subfloat[Whiteboard wiping\label{fig:tasks_wiping}]{%
        \includegraphics[width=0.4\linewidth]
            {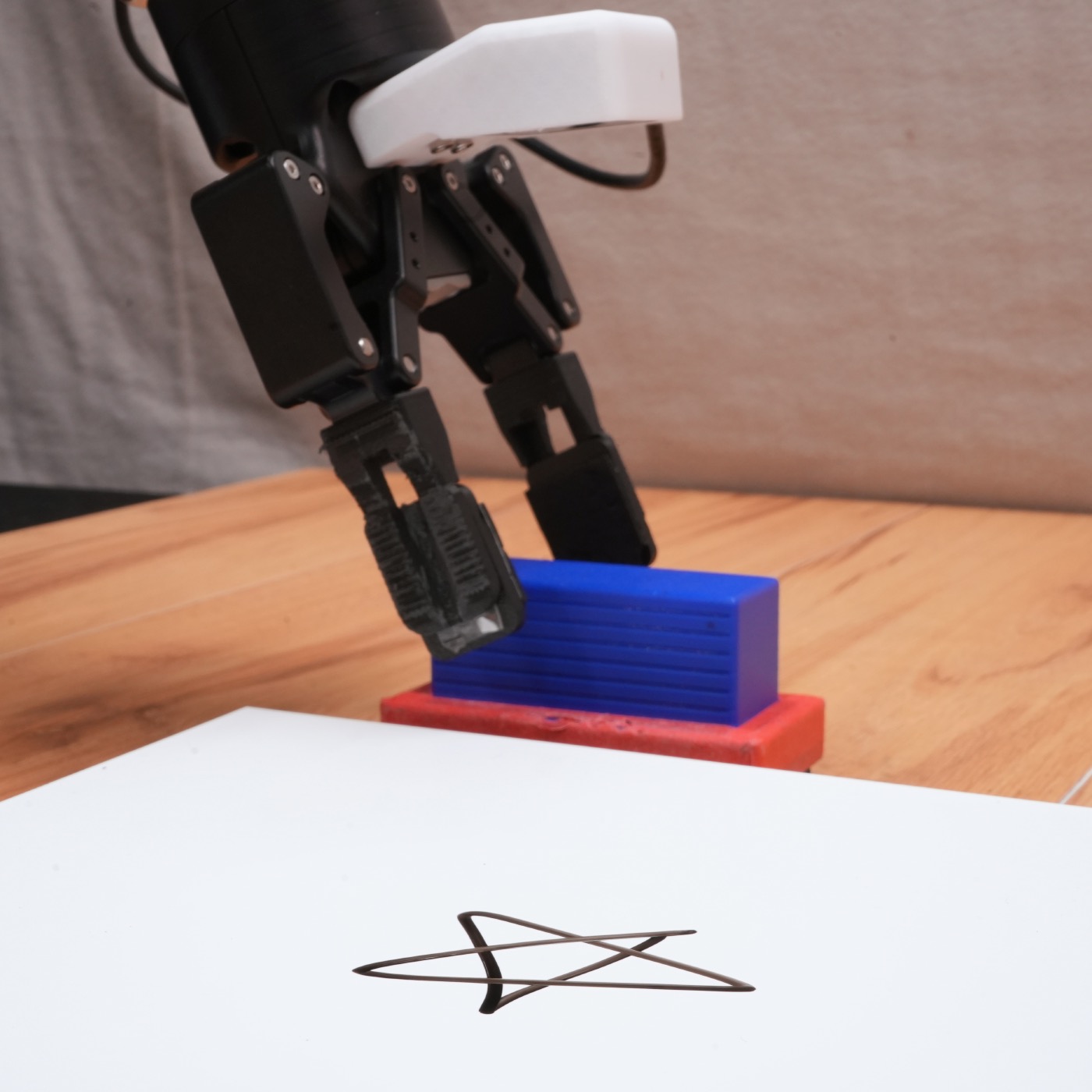}%
    }\hfill
    \subfloat[Gear insertion\label{fig:tasks_gear}]{%
        \includegraphics[width=0.4\linewidth]
            {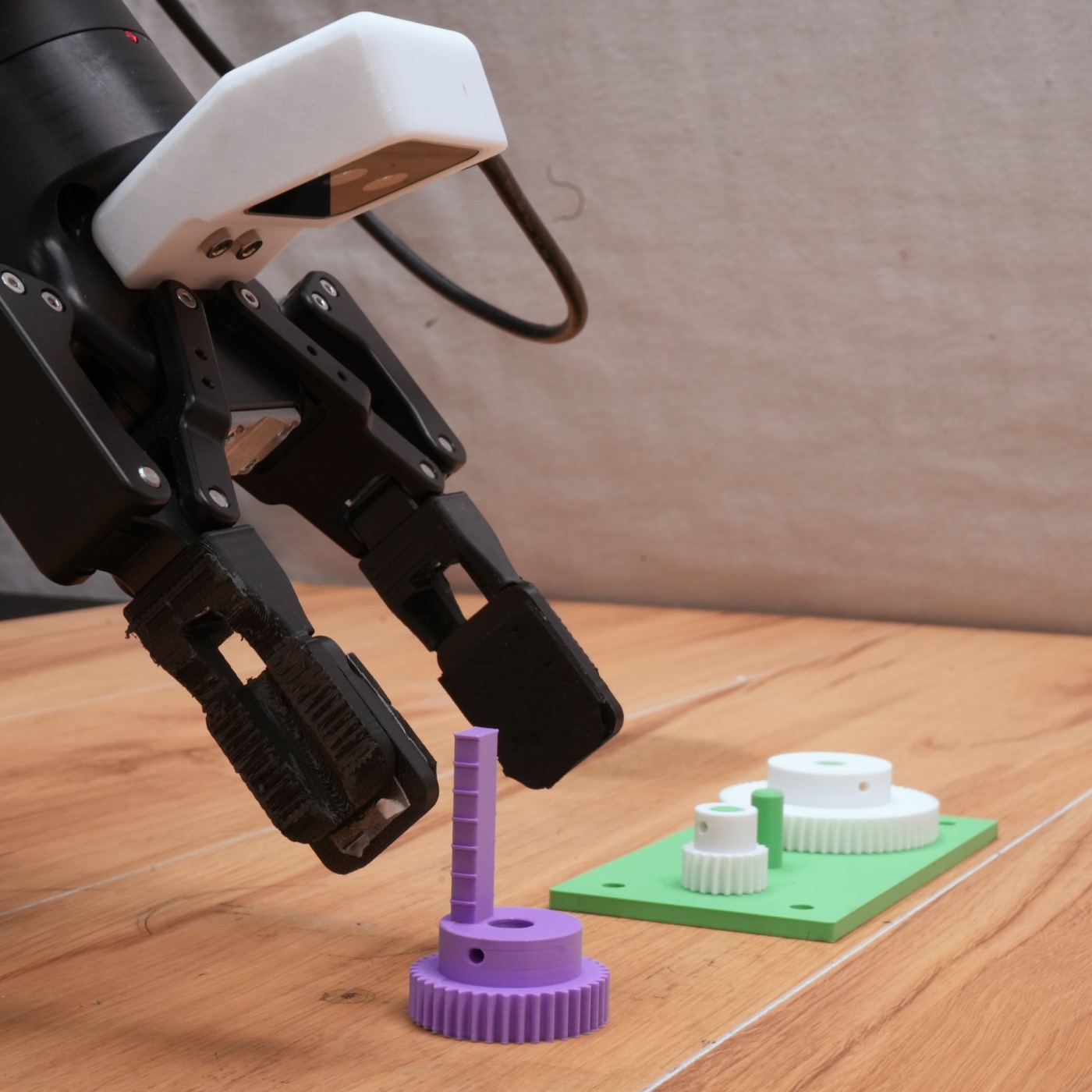}%
    }

    \par\vspace{2mm}

    \subfloat[Screw tightening\label{fig:tasks_screw}]{%
        \includegraphics[width=0.4\linewidth]
            {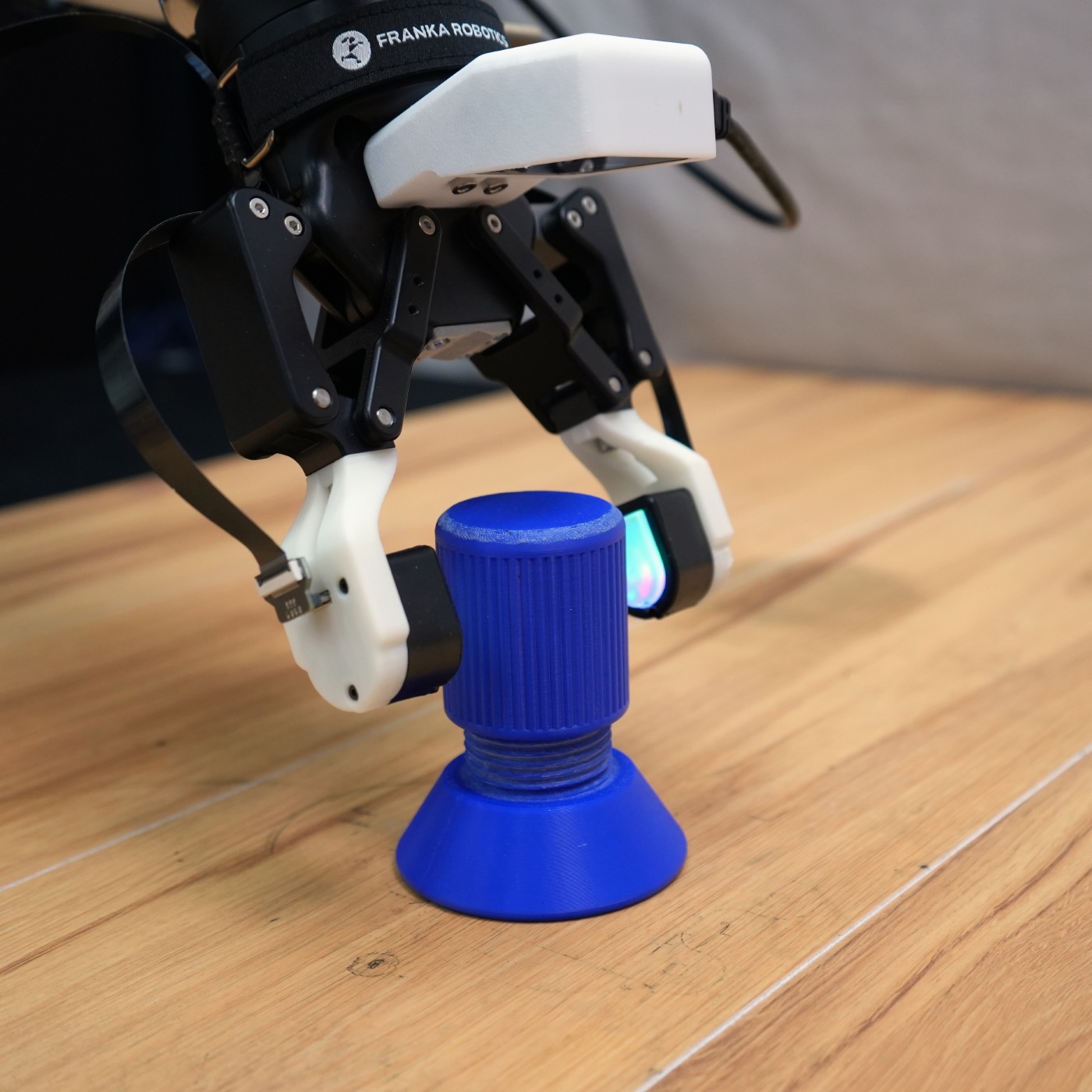}%
    }\hfill
    \subfloat[Box sorting\label{fig:tasks_sorting}]{%
        \includegraphics[width=0.4\linewidth]
            {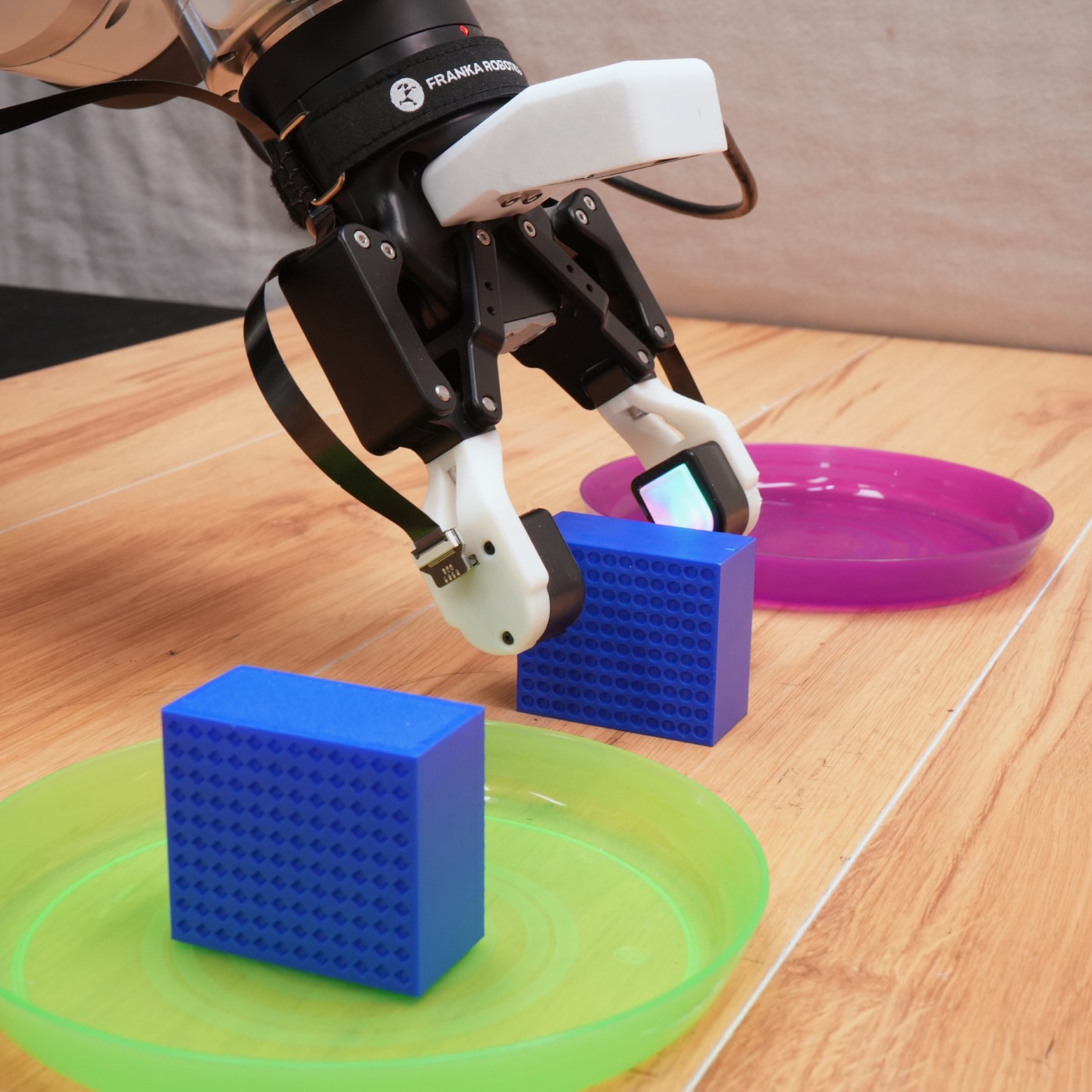}%
    }

    \caption{Real-world tasks used to evaluate our data
    acquisition approach: (a) whiteboard wiping,
    (b) gear insertion, (c) screw tightening, and
    (d) box sorting. Evaluation also includes one
    simulated task (not shown).}
    \label{fig:tasks}
\end{figure}
\begin{figure}[t]
    \centering
    \captionsetup[subfloat]{
        farskip=0pt,nearskip=0pt,captionskip=2pt
    }
    \vspace*{3.5mm}
    \subfloat[In-the-wild data collection\label{fig:setup}]{%
        \includegraphics[width=0.48\linewidth]
        {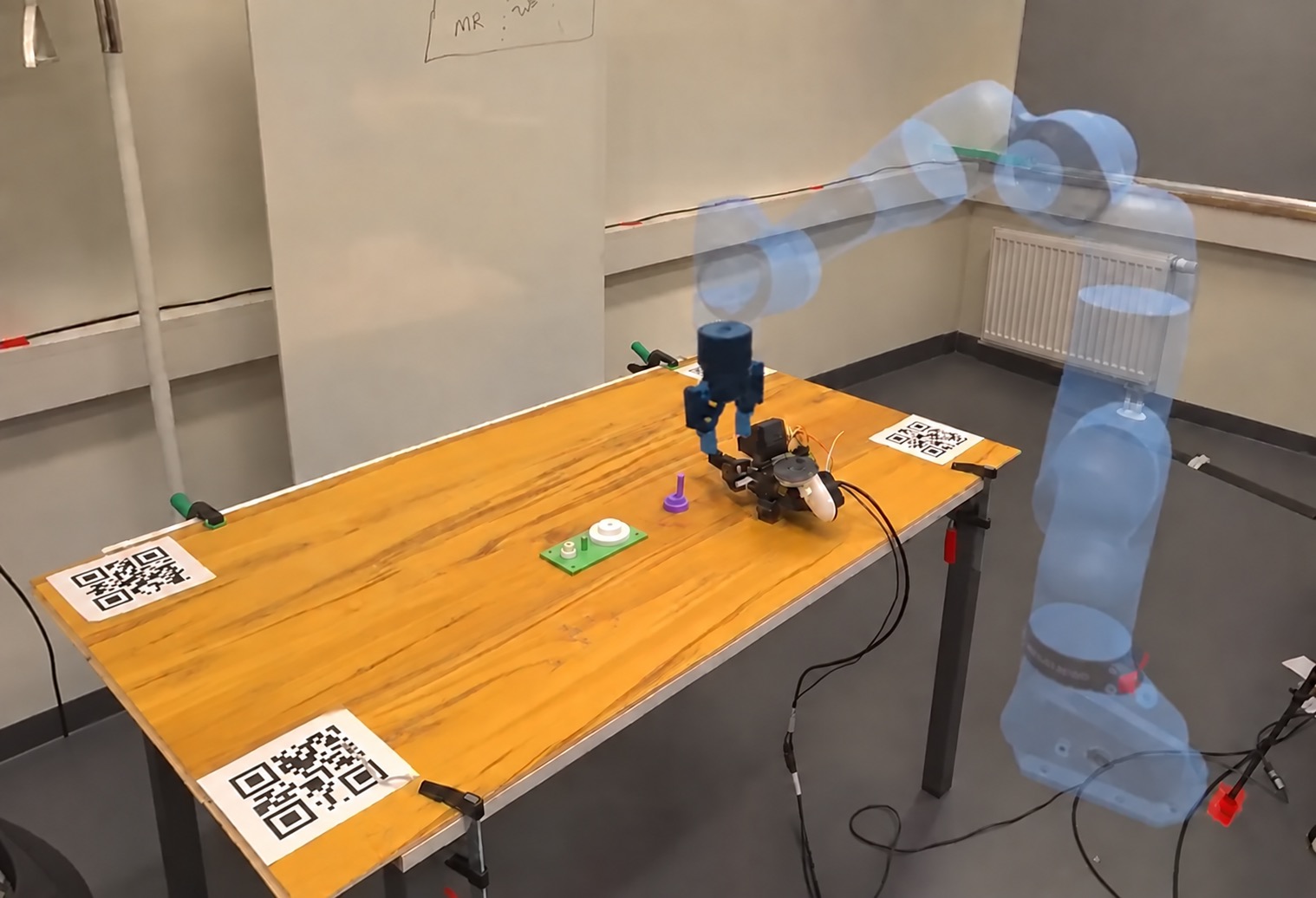}%
    }\hfill
    \subfloat[Simulated packing\label{fig:simulated_packing}]{%
        \includegraphics[width=0.48\linewidth]
        {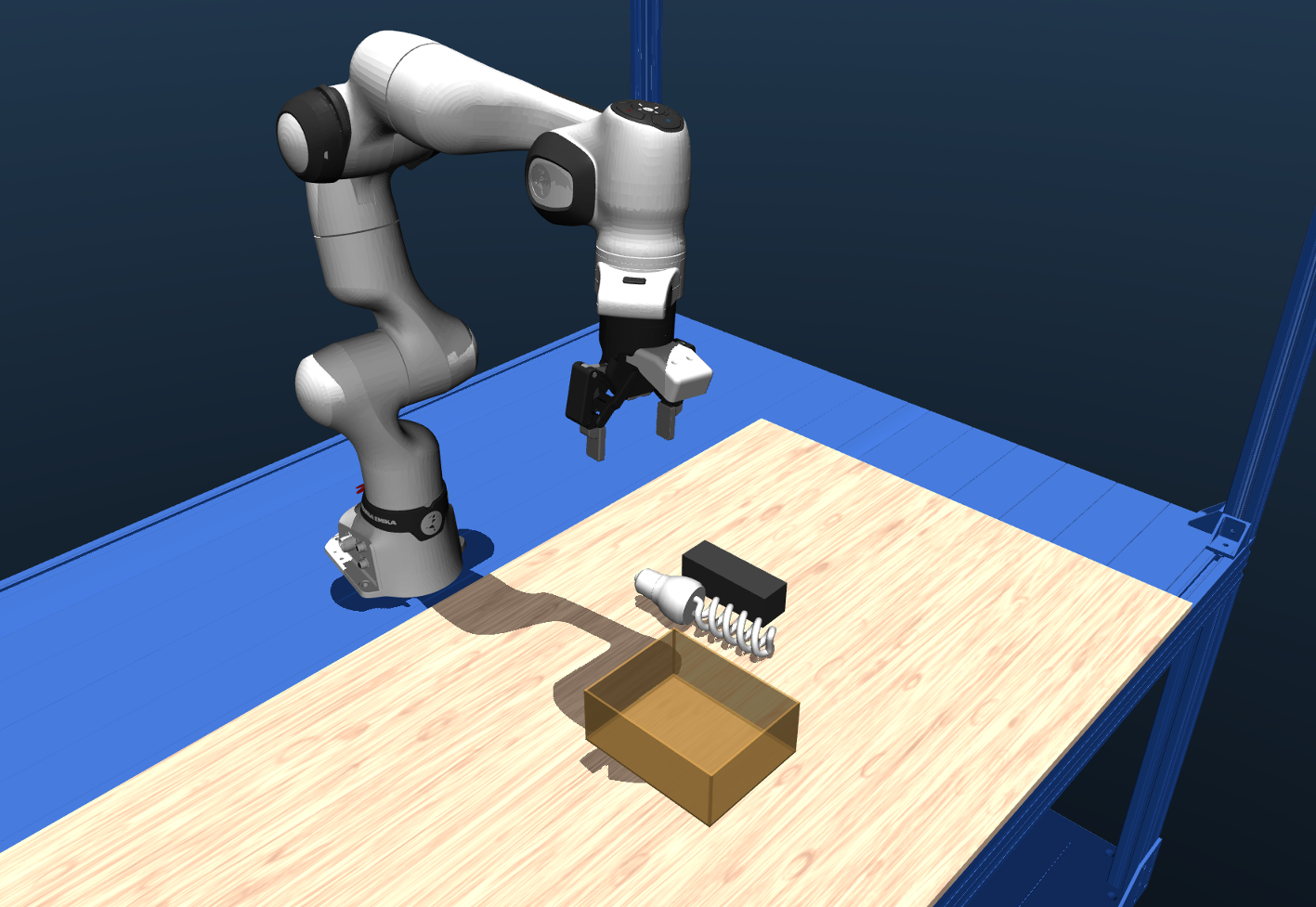}%
    }\\ 
    \vspace{3mm}
    \subfloat[Inference setup\label{fig:real_setup}]{%
        \includegraphics[width=0.48\linewidth]
        {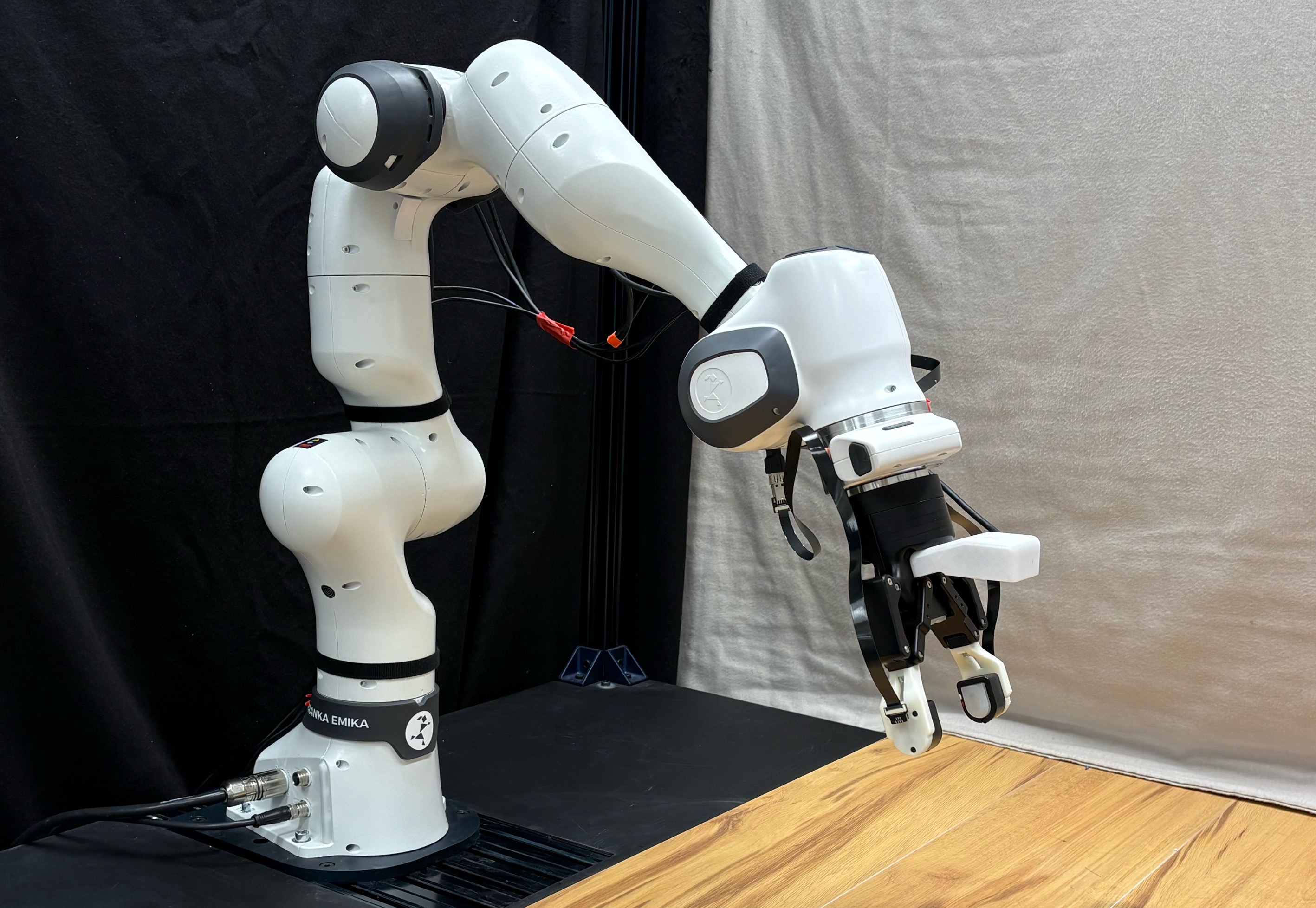}%
    }
    \caption{Experimental environments used to evaluate RUMI.
    (a) Robot-free data collection with QR-code markers and an augmented-reality robot overlay for recording-time validation.
    (b) Simulated packing task involving a fragile light bulb and a heavy transformer block.
    (c) Real-world setup for policy deployment using a Franka FR3 robot equipped with a Robotiq 2F-85 gripper.}
    \label{fig:environments}
\end{figure}

To validate the effectiveness of RUMI, we conduct a series of experiments, focusing on A) usability, B) transfer learning on an imitation policy, and C) haptic feedback utility. For experiments conducted in subsections A) and B), we set up an office table covered with wooden boards (Fig.~\ref{fig:setup}) to mimic the inference setup (Fig.~\ref{fig:real_setup}). We place one QR code at each table corner, and calibrate the base pose of the robot. Our inference environment consists of a Franka FR3 robot arm equipped with a Robotiq 2F-85 gripper and a RealSense D405 wrist camera. The haptic feedback utility experiment is conducted in a digital twin of the inference setup. The complete setup is shown in Fig.~\ref{fig:real_setup}.

\subsection{Data quality comparison }
To test whether our proposed AR-based digital twin projection provides meaningful assistance in the quality of demonstration data, we conduct a small user study. We recruit 5 participants with limited experience in teleoperating a robotic arm with VR controllers. The participants are asked to perform a simple box pick-and-place task. We collect demonstrations under two conditions. In the first condition, the participants perform the task without any sort of AR projection, and are informed beforehand to start each episode from the same starting pose as close as possible. We name this condition \textbf{w/o AR}. In the second condition, we enable the AR projection and the catch-to-teleop functionality. We call this condition \textbf{AR}. Each participant performs 10 demonstrations per condition, resulting in 100 total episodes. 

\begin{figure}
\vspace{5.1pt}
    \centering
    \includegraphics[width=0.95\linewidth]{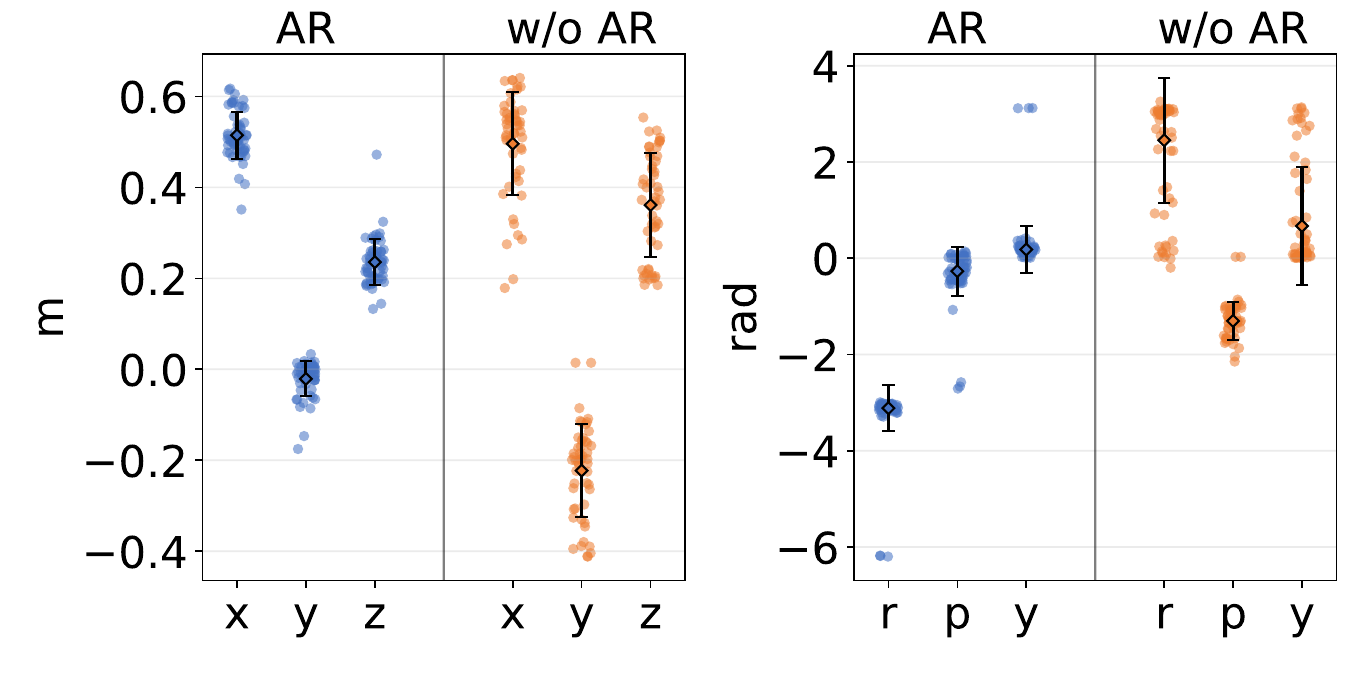}
    \includegraphics[width=0.95\linewidth]{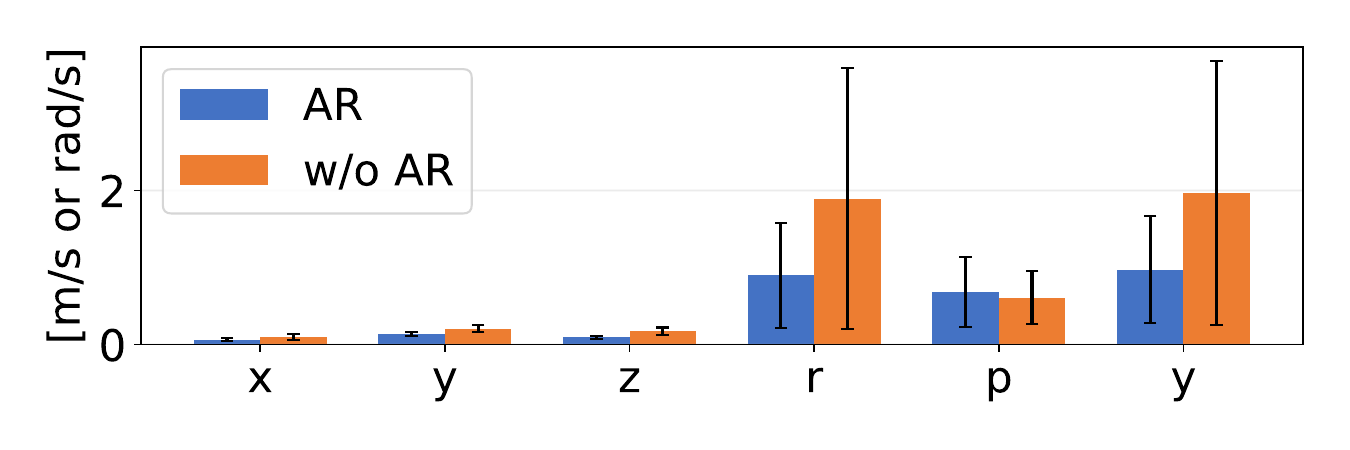}
    \caption{Top: Initial poses for all episodes under the two user study conditions, split across translation (left) and rotation (right). Bottom: Average Cartesian space velocity under the two conditions.}
    \label{fig:userstudy}
\end{figure}

With the collected data, we calculate 4 metrics: initial pose variance, average Cartesian space velocity of the VR controller, number of episodes containing joint limit and velocity violations, and average Yoshikawa manipulability across the whole dataset. Joint velocity and position limits are taken from FR3's technical data. The results of the analysis are shown in ~\autoref{fig:userstudy} and ~\autoref{tab:userstudy}. Looking at ~\autoref{tab:userstudy}, the AR condition results in fewer number of joint limit- and velocity violations. The average manipulability is 19.6\% higher than the w/o AR condition, though the standard deviation is also greater. \autoref{fig:userstudy} shows that the catch-to-teleop feature in the AR condition results in a much tighter cluster of initial poses. Additionally, the average Cartesian space velocity is lower, indicating that the projected robot encourages the participant to slow down until the digital shadow is able to catch up. Given that the participants are unfamiliar with VR-based robot teleoperation, the results suggest that our AR-based robot visualization can assist inexperienced users with collecting high-quality data.

\subsection{Imitation Learning from RUMI Demonstrations}

The second experiment is to test whether RUMI is capable of collecting a high-quality dataset that can be used for training an imitation policy on a real robot. Furthermore, to test RUMI's versatility in using different fingertip sensors, we collect datasets for the CoinFT~\cite{coinft} and the DIGIT~\cite{digit} optical tactile sensor. We design four contact-rich manipulation tasks, two for each sensor type, as shown in \autoref{fig:tasks}. Additionally, we divide each task into 3 broad stages: 1) grasp success, 2) task attempt, and 3) task success, in order to better assess where each policy variant fails. We use the Action Chunking Transformer (ACT)~\cite{zhao2023act} for all trained models.

For the CoinFT sensor, we collected datasets for board wiping and gear insertion tasks (\autoref{fig:tasks}a,b). %
In the board wiping task, the robot must first pick up a whiteboard eraser, which is fitted into a 3D-printed holder for easier grasp. Then, it must erase the star pattern drawn on the board positioned in front of it. The position and shape of the star pattern is varied across each trial.
The gear insertion task involves a small 3D-printed setup consisting of three gears, drawn from the ManipulationNet benchmark~\cite{chen2026manipulationnet}. The robot must pick up a gear and insert it into the center position of the gear assembly and ensure that the teeth align for a full insertion. The positions of the main assembly plate and the gear are not randomized. 

We evaluate the screw tightening and box sorting tasks for the DIGIT sensor, as the optical tactile sensor's high-resolution tactile information makes it suitable for sensing both shear force and surface texture (\autoref{fig:tasks}c,d). %
For the screw tightening task, the robot must tighten a cap already mounted on a 3D-printed screw fixture. The number of rotations required for tightening varies across each demonstration. 
The box sorting task involves two 3D-printed boxes with different surface texture patterns. The robot must then sort these boxes into their respective plates positioned adjacent to them.

We collect around 100 demonstrations per task with their appropriate fingertip sensors attached to the RUMI in the setup described above. For the CoinFT tasks, we train one model with force-adjusted impedance targets:
\begin{equation*}
    z' = z + F_{z,\text{CoinFT}}/K,
\end{equation*}

where $z$ is the recorded z-axis translation, $K$ is an empirically chosen value of \num{1000}, corresponding to \SI{10}{N/cm}, and $F_{z,\text{CoinFT}}$ is the z-axis force measured by the CoinFT sensor.
We train another model using the raw observed trajectories, with wrist camera images being available for both. For the DIGIT tasks, we follow the training recipe described in VITaL~\cite{vital}: we first perform contrastive pretraining on a ResNet-18-based~\cite{he2016deep} tactile encoder backbone along with a ResNet-18-based vision encoder backbone. This combined backbone is then frozen to train the corresponding ACT policy with tactile images. Again, another variant is trained without the tactile image input using just the observed trajectories and the wrist camera images. This setup allows us to test whether tactile data collected with RUMI contribute meaningfully towards completing the tasks successfully. We evaluate each model variant 20 times. The results are shown in \autoref{tab:task_success}.

The results show a clear correlation between task success rates and the availability of their respective fingertip tactile sensors. In the gear assembly task, we observe that the force-adjusted trajectories are able to reliably push the gear into alignment with its neighbors, also aided by the force-adjusted gripper strength that allows for passive compliance. In the no-tactile case, although the gear is often successfully inserted through the stick, the lack of further downward trajectory means that the policy often releases its grasp without pushing the gear into alignment. For the board task, the no-tactile variant is not able to completely wipe the pattern off the whiteboard despite passing over it with the eraser due to the lack of downward pressure. 

The majority of failures with the screw tightening DIGIT variant comes from the fact that the policy struggles to maintain a stable contact with the cap for long enough to fully tighten it, resulting in slippage. This is expected, as RUMI operation creates a lot of jitter in the demonstration, which makes the contact stage of the task unstable. Nonetheless, the DIGIT variant outperforms the raw variant, which frequently released its grasp after barely rotating the cap, as it lacks the shear force signal to understand when the task is completed. For the box sorting task, the raw variant overfits to one particular trajectory, as it lacks information about the different boxes' textures. The DIGIT variant was able to successfully sort the box correctly in all but one valid attempt. These results indicate that RUMI is capable of recording useful demonstrations with different fingertip sensors.

\begin{table}[t]
    \centering
    \vspace{5.1pt}
    \caption{Ablation of data quality with and without AR projection. Joint limit- and velocity violations report affected episodes/total episodes, and average manipulability indicates the average value across the entire dataset.}
    \label{tab:userstudy}
    \renewcommand{\arraystretch}{1.2}
    \setlength{\tabcolsep}{12pt}
    \begin{tabular}{lcc}
        \toprule
        \textbf{Metric}
        & \textbf{AR}
        & \textbf{w/o AR} \\
        \midrule
        Joint Limit Vio. $\downarrow$
        & \textbf{3/50}
        & 16/50 \\
        Joint Velocity Vio. $\downarrow$
        & \textbf{11/50}
        & 29/50 \\
        Avg. Manip. $\uparrow$
        & \textbf{$0.055 \pm 0.022$}
        & $0.046 \pm 0.0057$ \\
        \bottomrule
    \end{tabular}
\end{table}

\subsection{Evaluation in Simulation}
We investigate whether RUMI can collect force-aware demonstrations in simulation and whether the contact-triggered blocker helps operators to provide object-dependent grasp-force commands. Using the MuJoCo simulator, the experiment reproduces an industrial packing task of transferring a fragile light bulb~\cite{lightbulb} and a substantially heavier transformer block into a shipping box. 
Here, we set the gripper force amplification factor to 1.5 to reduce the force required from the operator. The two objects require distinct handling strategies: the bulb must be grasped gently, whereas the transformer requires a stronger grasp to prevent slip.

We collect around 100 demonstrations in total and use them to train the same imitation policy described in the previous subsection. A rollout is successful when the selected object is transferred into the shipping box. Bulb rollouts additionally fail if the maximum grasp force exceeds \SI{10}{N}, which serves as a simulated fracture threshold. For the transformer, insufficient grasp force leads to object slippage.

To isolate the effect of haptic contact feedback during data collection, we train a second policy using demonstrations recorded in the same setting with the contact trigger and blocker disabled. In this condition, operators command the gripper through a position controller, where grasp force arises implicitly from the position error. The operator then visually estimates the quality of the grasp and adjusts the gripper width as needed. We compare the mean grasp force in the demonstrations (F$_\text{DS}$), the mean force produced by policy rollouts (F$_\text{R}$), and the success rate  based on 100 rollouts for each object and feedback condition.

As shown in Table~\ref{tab:blocker_ablation}, demonstrations collected with haptic feedback exhibit a clear object-dependent force strategy: operators apply \SI{5.66}{N} to the bulb and \SI{22.82}{N} to the transformer on average. The learned policies preserve this separation during rollout, producing mean forces of \SI{5.42}{N} and \SI{19.86}{N}, respectively, and achieve success rates of 94\% and 97\%. Without haptic feedback, the demonstration forces converge toward less task-appropriate values: bulb force rises to \SI{18.46}{N}, causing every rollout to exceed the fracture threshold, while transformer force falls to \SI{15.83}{N}. These results indicate that the blocker provides a useful contact reference for demonstrating force-sensitive behavior and that the learned policy reproduces the resulting object-dependent grasp strategy.

\begin{table}[t]
\vspace{5.1pt}
    \centering
    \caption{Task success rates across individual stages and overall execution. Bracket indicates the sensor used to train the model.}
    \label{tab:task_success}
    \begin{tabular}{lccc}
        \toprule
        \textbf{Task} 
        & \textbf{Grasp} 
        & \textbf{Attempt} 
        & \textbf{Success} \\
        \midrule
        Gear Assembly (CoinFT) & 20/20 & 20/20 & 20/20 \\
        Gear Assembly (None) & 20/20 & 20/20 & 7/20 \\
        Whiteboard Wiping (CoinFT) & 20/20 & 18/20 & 16/20 \\
        Whiteboard Wiping (None) & 20/20 & 17/20 & 7/20 \\
        Screw Tightening (DIGIT)         & 17/20 & 17/20 & 9/20 \\
        Screw Tightening (None)         & 17/20 & 17/20 & 3/20  \\
        Box Sorting (DIGIT)        & 18/20    & 18/20    & 17/20    \\
        Box Sorting (None)        & 19/20    & 19/20    & 9/20    \\
        \bottomrule
    \end{tabular}
\end{table}

\section{Conclusions}
In this work, we presented \methodeName, a gripper-matched handheld interface for tactile-enabled data collection, modeled after the Robotiq 2F-85 gripper. The device combines gripper-pose tracking with compatibility to diverse fingertip sensors and user-side grasp-force measurement and provides haptic feedback during teleoperation. 
We demonstrated its usability for robot-free data collection, imitation policy training, and simulated teleoperation.
By providing a consistent physical interface across these modes, \methodeName~enables the collection of embodiment-matched, tactile-aware demonstrations for robot learning. 
Our method's current limitations include being restricted to only a wrist camera viewpoint and the lack of external force feedback to complement the blocking mechanism.
Future work could include generative AI to overcome the single-view limitation as well as more informative visual cues about external forces acting upon the simulated or teleoperated robot.

\section*{Acknowledgment}
\begin{table}[t]
\vspace{5.1pt}
    \centering
    \caption{Simulated packing results. Haptic feedback produces object-dependent grasp forces in demonstrations (F$_\text{DS}$) that persist in policy rollouts (F$_\text{R}$).}
    \label{tab:blocker_ablation}
    \begin{tabular}{lccc}
        \toprule
        \textbf{Packing Task} 
        & \textbf{Avg. F$_\text{DS}$ [N]} 
        & \textbf{Avg. F$_\text{R}$ [N]}
        & \textbf{Success ($\uparrow$)} \\
        \midrule
         Bulb  & 5.66 & 5.42 & 94\% \\
        Block  & 22.82 & 19.86 & 97\% \\
         Bulb (no haptic) & 18.46 & -- & 0\% \\
        Block (no haptic) & 15.83 & 13.67 & 99\% \\
        \bottomrule
    \end{tabular}
\end{table}
\blackout{All work related to the CoinFT sensors was performed by and at UTN. Siemens' contribution was limited to the mechanical design of the device. The authors gratefully acknowledge the scientific support and HPC resources provided by the Erlangen National High Performance Computing Center (NHR@FAU) of the Friedrich-Alexander-Universität Erlangen-Nürnberg (FAU) under the BayernKI project v106be. BayernKI funding is provided by Bavarian state authorities. This work has been partially supported by the German Federal Ministry of Research, Technology and Space (BMFTR) under the Robotics Institute Germany (RIG). This work has been partially supported by the project GeniusRobot funded by the German Federal Ministry of Education and Research (BMBF grant no. 01IS24083). We further thank Volker Schneider and Devadas Sheela for their assistance with hardware development and setup.}
OpenAI ChatGPT and Claude Code were used to assist with debugging and the generation of limited boilerplate and data-processing code. All generated code was reviewed, adapted, and validated by the authors. The research methodology, algorithm design, analysis, and interpretation of results were performed by the authors. ChatGPT was further used to remove background elements from Figure \ref{fig:environments} a.

\bibliographystyle{IEEEtran}
\bibliography{sources.bib}

\end{document}